\documentclass{article}

\usepackage[preprint]{neurips_2024}

\usepackage{mathtools,amsfonts,amsmath,amssymb}
\usepackage{bm}

\def\vx{{\bm{x}}}

\def\vz{{\bm{z}}}

\def\mG{{\bm{G}}}

\def\mX{{\bm{X}}}

\def\mZ{{\bm{Z}}}

\DeclareMathAlphabet{\mathsfit}{\encodingdefault}{\sfdefault}{m}{sl}
\SetMathAlphabet{\mathsfit}{bold}{\encodingdefault}{\sfdefault}{bx}{n}

\def\gL{{\mathcal{L}}}

\def\gT{{\mathcal{T}}}

\def\1{\bm{1}}

\def\Indic{{\bm{1}}}

\newcommand{\R}{\mathbb{R}}

\def\floor#1{\left\lfloor #1 \right\rfloor}

\def\paren#1{\left( #1 \right)}

\def\norm#1{\left\| #1 \right\|}

\DeclareMathOperator{\ssl}{ssl}
\usepackage{amsthm}
\newtheorem{theorem}{Theorem}

\usepackage[utf8]{inputenc} 
\usepackage[T1]{fontenc}    
\usepackage[hidelinks, colorlinks=true, citecolor=NavyBlue]{hyperref}       
\usepackage{url}            
\usepackage{booktabs}       
\usepackage{amsfonts}       
\usepackage{nicefrac}       
\usepackage{microtype}      
\usepackage[dvipsnames]{xcolor}
\usepackage{graphicx}
\usepackage{cleveref}
\usepackage{makecell}
\usepackage{subcaption}
\usepackage{amsmath}
\usepackage{bbm}
\usepackage{enumitem,amssymb}
\usepackage{wrapfig}

\newcommand{\xclr}{$\mathbb{X}$-CLR}

\newcommand{\Mark}[1]{}
\newcommand{\Vlad}[1]{}
\newcommand{\Vivien}[1]{}
\newcommand{\Pietro}[1]{}
\newcommand{\Diane}[1]{}
\newcommand{\Randall}[1]{}
\newcommand{\Cho}[1]{}
\newcommand{\Yann}[1]{}

\title{$\mathbb{X}$-Sample Contrastive Loss: Improving Contrastive Learning with Sample Similarity Graphs}

\author{
  Vlad Sobal$^{1,2}$ \quad Mark Ibrahim$^{1}$ \quad Randall Balestriero$^{3}$\quad Vivien Cabannes$^{1}$ \\ 
  \textbf{Diane Bouchacourt}$^{1}$ \quad \textbf{Pietro Astolfi}$^{1}$ \quad \textbf{Kyunghyun Cho}$^{2,4,5}$ \quad \textbf{Yann LeCun}$^{1,2}$ \\
  $^1$Meta FAIR \quad $^2$New York Univeristy \quad $^3$Brown University \quad $^4$Genentech \quad $^5$CIFAR \\
}

\begin{document}

\maketitle

\begin{abstract}
Learning good representations involves capturing the diverse ways in which data samples relate. Contrastive loss—an objective matching related samples—underlies methods from self-supervised to multimodal learning. Contrastive losses, however, can be viewed more broadly as modifying a similarity graph to indicate how samples should relate in the embedding space.  This view reveals a shortcoming in contrastive learning: the similarity graph is binary, as only one sample is the related positive sample. Crucially, similarities \textit{across} samples are ignored. 
Based on this observation, we revise the standard contrastive loss to explicitly encode how a sample relates to others. We experiment with this new objective, called $\mathbb{X}$-Sample Contrastive, to train vision models based on similarities in class or text caption descriptions.
Our study spans three scales: ImageNet-1k with 1 million, CC3M with 3 million, and CC12M with 12 million samples. The representations learned via our objective outperform both contrastive self-supervised and vision-language models trained on the same data across a range of tasks. When training on CC12M, we outperform CLIP by $0.6\%$ on both ImageNet and ImageNet Real. Our objective appears to work particularly well in lower-data regimes, with gains over CLIP of $17.2\%$ on ImageNet and $18.0\%$ on ImageNet Real when training with CC3M. Finally, our objective seems to encourage the model to learn representations that separate objects from their attributes and backgrounds, with gains of $3.3$-$5.6$\% over CLIP on ImageNet9. We hope the proposed solution takes a small step towards developing richer learning objectives for understanding sample relations in foundation models.
\end{abstract}

\section{Introduction}

\begin{figure}
\centering
\begin{subfigure}[t]{0.57\linewidth}
\centering
\includegraphics[width=\textwidth]{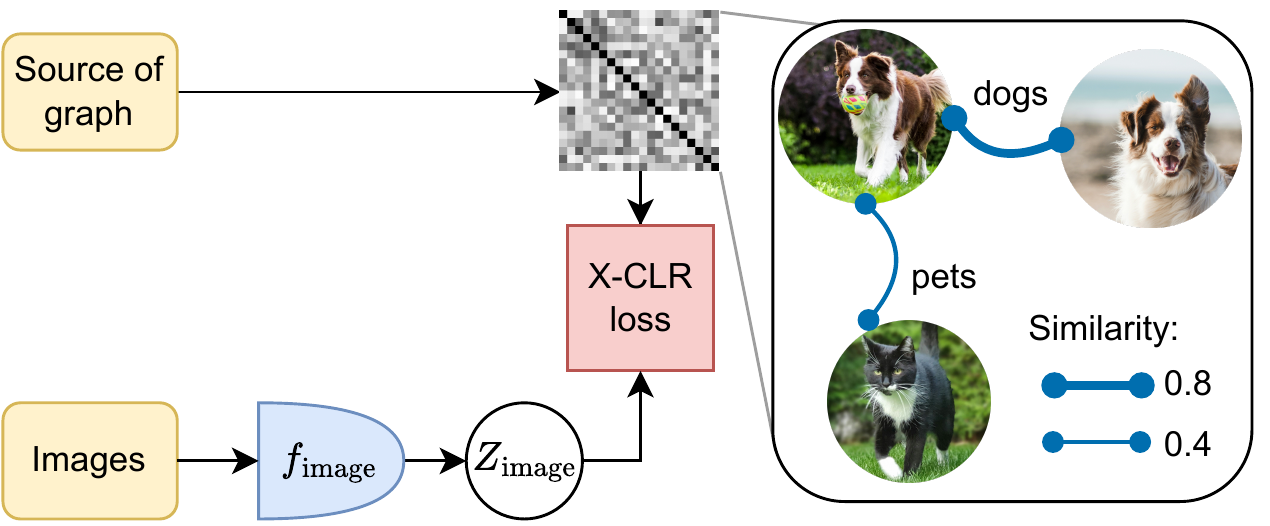}
\caption{}
\label{fig:fig1_new}
\end{subfigure}%
\hspace{0.01\linewidth}%
\begin{subfigure}[t]{0.40\linewidth}
\centering
\includegraphics[width=\textwidth]{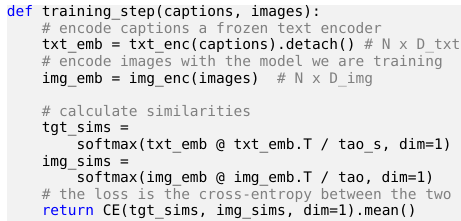}
\caption{}
\label{fig:code_listing}
\end{subfigure}
\label{fig:idea}
\caption{ \textbf{a) The diagram of \xclr{}}. \xclr{} objective learns representations of images with the help of a soft relationship graph. The graph can be built based on accompanying data, e.g. taxonomy for biological data. In our experiments, we use captioned images, and build similarities based on the similarity of captions. \textbf{b) Python-style pseudo-code of \xclr{} with similarity based on text captions.}}
\label{fig:overall_figure}
\end{figure}


Contrastive loss underlies methods from self-supervised learning (SSL) to multimodal learning \citep{radford2021learning, chen2020simple, oord2018representation}. In SSL, contrastive learning encourages the model to associate a sample with another view of the sample created using hand-crafted data augmentation—this related view is the positive sample. Other samples are then pushed away as negative, unrelated samples in the models’ representation space. Contrastive losses also play a crucial role in multimodal models such as CLIP \citep{radford2021learning}, where the model associates an image with its text caption in representation space. Here contrastive learning designates the caption and image representations as positives while all other text-image pairs are designated as unrelated negatives.

More broadly, contrastive losses can be seen as modifying a similarity graph to indicate how samples should relate in the model’s representation space \citep{cabannes2023active}. This view reveals a shortcoming in contrastive learning: the similarity graph is binary, as only one sample is the related positive sample. Crucially, similarities across samples, containing precious signals about how aspects of one sample may relate to another, are ignored. For example, as shown in \cref{fig:idea}, contrastive learning treats each text-image pair independently, without explicitly encoding similarities in the images depicting dogs and the others sharing a grassy background. Standard contrastive objectives do not explicitly account for similarities across samples, thereby limiting the quality of the learned representations. Here, we explore here how to capture such similarities by modifying the standard contrastive objective.

To account for similarities across samples, we first remove the binary negative vs. positive designations in standard contrastive loss. We introduce instead a similarity graph with continuous scalars capturing the extent to which two samples are related. Consider the example in \cref{fig:idea}, where the two dog images have a high similarity while the dog and cat images have a more moderate similarity. We experiment with this new objective, called $\mathbb{X}$-Sample Contrastive ($\mathbb{X}$-CLR), by training vision models using a graph of similarities inferred from class or text caption descriptions found in common datasets. Our study spans three training dataset scales from 1 million samples with high-quality labels from ImageNet \citep{imagenet} to 3 and 12 million noisy image-text caption pairs from CC3M and CC12M \citep{sharma2018conceptual}. 

We find that compared to contrastive baseline methods trained on the same data, representation trained using $\mathbb{X}$-CLR outperform contrastive training on a range of tasks from standard classification to tasks involving the decomposition of objects from their attributes and backgrounds. When training on CC12M, we outperform CLIP by 0.6\% on both ImageNet and ImageNet Real \citep{beyer2020we}. Furthermore, $\mathbb{X}$-CLR seems to encourage the model to learn representations that separate objects from their attributes and backgrounds, with gains of 3.4-4.9\% over CLIP on ImageNet9 \citep{xiao2020noise}. 
We also find for fine-grained disambiguation of object attributes, the quality of labels used to infer the similarity graph is much more important than the data quantity. Compared to noisier web caption data, we find $\mathbb{X}$-CLR trained on 1 million higher quality class labels outperforms representations learned via standard contrastive CLIP trained $12\times$ more data. Finally, we find $\mathbb{X}$-CLR appears to work particularly well in lower-data regimes, with gains over CLIP of 16.8\% on ImageNet and 18.1\% on ImageNet Real when training with CC3M. In short, we find representations learned using $\mathbb{X}$-CLR generalize better, decompose objects from their attributes and backgrounds, and are more data-efficient.

Our contributions are:

\begin{enumerate}
    \item  We present a graph similarity perspective of contrastive losses, revealing standard losses encode a sparse similarity matrix that treats other, related, samples as negatives.
    \item  Consequently, we propose a new $\mathbb{X}$-CLR loss that explicitly accounts for similarities across samples
    \item  We experiment with this objective across three levels of data scale from 1-12 million samples.
    \item  We find representations learned via $\mathbb{X}$-CLR
    \begin{enumerate}
        \item  Generalize better on standard classification tasks with consistent gains over contrastive baselines trained on the same data. For example, when training on CC12M we outperform CLIP by 0.6\% on both ImageNet and ImageNet Real.
        \item  Disambiguate aspects of images such as attributes and backgrounds more reliably, with gains of 3.3-5.6\% over CLIP on background robustness benchmarks for ImageNet.
        \item  Finally, we find $\mathbb{X}$-CLR learns more efficiently when data is scarce, with gains of 17.2\% on ImageNet and 18.0\% on ImageNet Real when pretraining on the smaller 3 million sample CC3M dataset.
    \end{enumerate}
\end{enumerate}

We hope the proposed solution takes a small step towards developing richer learning objectives for understanding sample relations in foundation models to encode richer, more generalizable representations.

\section{Related Work}

\paragraph{Contrastive learning} Various contrastive objectives have been proposed over the years \citep{contrastive, schroff2015facenet}. More recently, the InfoNCE objective \citep{oord2018representation} has been the most popular choice for self-supervised methods, e.g. SimCLR \citep{chen2020simple} and MoCo \citep{he2020momentum}. InfoNCE objective has also been successfully used to learn vision-language models using CLIP \citep{radford2021learning}.
The basis of those objectives is to make positive pairs have similar representations,
while the negatives, which typically are just all other elements in a batch, should have a different representation. In its original form, InfoNCE is binary, meaning it only works with positive and negative pairs, and does not support degrees of similarity. The positive pairs are usually two augmentations of the same sample, which makes well-tuned augmentations crucial for good performance \citep{ryali2021characterizing}. \citet{dwibedi2021little} estimate positives using nearest neighbors in the latent space instead and therefore can use weaker augmentations, while \cite{caron2020unsupervised} use cluster assignment.
A few methods have proposed modifications wherein multiple positive pairs are supported,
e.g., \citet{khosla2020supervised} groups positive by class labels, \citet{hoffmann2022ranking} propose 
using WordNet \citep{wordnet} hierarchy to define ranked positive samples, and \citet{tian2024stablerep} uses a generative model to obtain multiple positives for the same concept.
\citet{haochen2021provable} also look at contrastive learning through
the lens of graphs, and propose a novel spectral objective.
\citet{wang2023message} draw connections between contrastive learning and message passing on the augmentation graph, while \citet{wang2022chaos} show that aggressive data-augmentations like cropping can connect samples of the same class.
\citet{zhang2023generalization} show that contrastive learning
objective implicitly learns the graph in which the samples are connected via augmentations in the case of SimCLR or via captions in the case of CLIP.
However, in that paradigm only visually similar samples or samples with a common caption get connected in the graph, while in our proposed method the samples are connected based on the semantics, and therefore visually dissimilar samples can be connected.

\paragraph{Soft targets} Using soft targets provides more learning signal to the model,
possibly making it learn better and faster. This has been explored with distillation by \citet{hinton2015distilling}.
Soft targets have also been used with InfoNCE in the context of distillation in ReSSL \citep{zheng2021ressl} and SCE \citep{denize2023similarity}, where the target cross-sample similarity comes from the teacher model. \citep{feng2023maskcon} use soft targets from self-distillation to train an image encoder with coarse labels. Similarly, \citet{fini2023semi} compute soft targets via latent clustering and apply it to semi-supervised learning. \citet{shen2023inter} use patch-mixing to train ViT image encoders to model inter-sample relationships.
\citet{andonian2022robust} proposes to use soft targets for CLIP \citep{radford2021learning} training, 
and calculates the targets via self-distillation. \citet{wu2023tinyclip} use a similar objective to ours to distill the CLIP model into a smaller one. Further soft CLIP objectives are explored by \citet{fini2023improved}, who apply label smoothing to obtain soft targets, and \citet{gao2024softclip}, who estimate soft targets by comparing fine-grained image information. Finally, \citet{huang2024cross} train CLIP with
non-zero cross-sample similarities computed based on pre-trained uni-modal models for
text and vision. In this study, we build on the work of \cite{cabannes2023active} who propose a unifying framework to view SSL and supervised learning objectives as learning with different underlying similarity graphs. We take inspiration from the soft targets literature and propose using a soft graph. As opposed to distillation, we focus more on the
graph design, and try different graph sources, including ones not based on distillation, see \cref{tab:similarities}.

\section{Understanding contrastive losses via similarity graphs}

\subsection{X-Sample Graphs}

\begin{figure}
\centering
\includegraphics[width=0.8\textwidth]{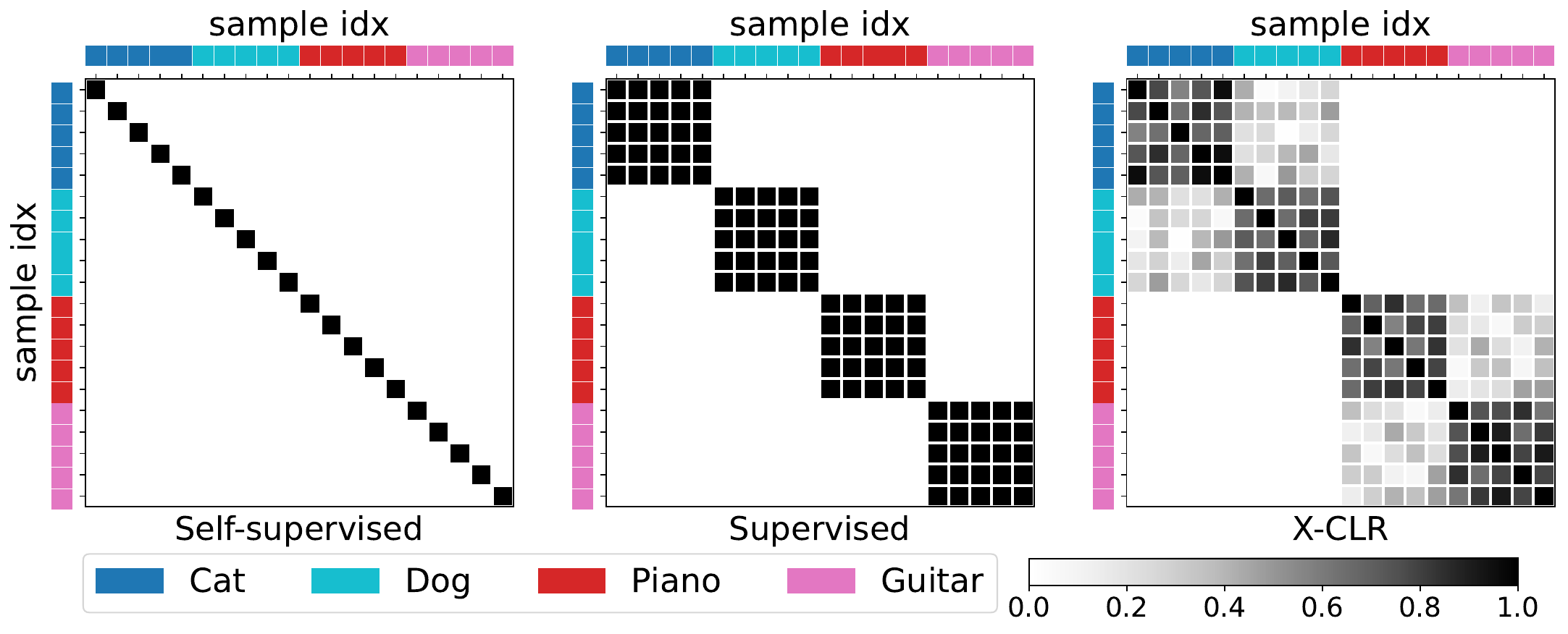}

\caption{\textbf{Sample similarity adjacency matrices of existing methods vs. our $\mathbb{X}$-Sample Contrastive similarity loss (right).} We show pairwise similarities of 20 samples belonging to 4 classes. Similarity of 1 means the samples are identical, 0 -- they are completely unrelated. In case of self-supervised learning, none of the inter-sample relationships are modelled (left). Supervised learning relies on the labels to group samples of the same class together (center). \xclr{} models inter-class relationships by associating cats with dogs and pianos with guitars.}
\label{fig:method}
\end{figure}

Throughout this study, a similarity graph denotes a graph in which the nodes represent data samples, and edges similarity -- relationships. Given the number of data samples in the dataset $N$, a graph is expressed through its symmetric adjacency matrix $\mG \in \R^{N \times N}$, the semantic relation between inputs $i$ and $j$ being encoded in the real entry $\mG_{i,j}$. In \cref{fig:method}, we show graphs of different learning paradigms.
SSL does not rely on labels, but on positive pairs/tuples/views generated at each epoch. 
Let us denote by $V$ the number of positive views generated, commonly $V=2$ for positive pairs, and denote by $E$ the training epochs. In that case, the original $N$ input samples are transformed into $N\times V \times E$ ``augmented'' samples
\begin{align*}
    \mX^{(A)} &\triangleq [\underbrace{\gT(\vx_1),\dots,\gT(\vx_1)}_{\text{repeated $V\times E$ times}} ,\dots, \gT(\vx_N),\dots,\gT(\vx_N)]^\top,
\end{align*}
where each $\gT$ is a random input transformation with its own random parameters. The corresponding graph is given by:
\begin{align}
\mG^{(\ssl)}_{i,j} = \Indic_{\{\floor{i/VE}=\floor{j/VE}\}},\label{eq:G_ssl}
\end{align}
where the associated similarity graph captures if two samples were generated as augmentations of the same original input.  Such graphs $\mG$, as defined by \cref{eq:G_ssl}, are the ones used as targets in common SSL methods, as formalized below denoting $\mZ \triangleq f_\theta(\mX) \in \R^{N\times K}$.

\begin{theorem}[\citep{cabannes2023active}]
\label{lemma:characterization}
VICReg \citep{bardes2021vicreg}, SimCLR \citep{chen2020simple}, and BarlowTwins \citep{zbontar2021barlow} losses can be expressed in terms of the graph $\mG$ \eqref{eq:G_ssl}
\begin{align*}
    \\ \gL_{\rm VIC^2}(\mZ;\mG)=&\| \mZ\mZ^T  - \mG \|_F^2,
    \\ \gL_{\rm SimCLR}(\mZ;\mG)=&-\hspace{-0.2cm}\sum_{i,j\in[N]}\mG_{i,j}\log\paren{\frac{\exp(\tilde\vz_i^\top \tilde\vz_j)}{\sum_{k\in[N]} \exp(\tilde\vz_i^\top \tilde\vz_k)}},
    \\ \gL_{\rm BT}(\mZ;\mG)=& \norm{\tilde\mZ^\top \mG \tilde\mZ - I}^2,
\end{align*}
where $\tilde\vz \triangleq \vz / \norm{\vz}$ and $\tilde\mZ$ the column normalized $\mZ$ so that each column has unit norm.
\end{theorem}

In our study, we will focus on contrastive learning, i.e., SimCLR family of losses. We will demonstrate how to move away from the ad-hoc graph $\mG$ from \cref{eq:G_ssl}.

\subsection{Revisiting contrastive losses with similarity graphs: $\mathbb{X}$-CLR}

We introduce the soft cross-sample similarity to the widely used InfoNCE objective \citep{oord2018representation}. We note that the proposed framework isn't necessarily
only limited to InfoNCE-based methods and can potentially be integrated into non-contrastive objectives such as BYOL \citep{grill2020bootstrap}, SimSiam \citep{chen2020exploring}, or VICReg \citep{bardes2021vicreg}, although we leave the extension to other objectives for future work. In SimCLR \citep{chen2020simple}, given a batch of $N_b$ images, each image is augmented twice, so each sample has a true positive. The $2N_b$ images are then encoded to get representation vectors $z$.
Then:
\begin{align*}
& p_{i,j}=\frac{\exp(\mathrm{sim}(z_i, z_j)/\tau)}{\sum_{i=1}^{2N} \mathbbm{1}_{[k \neq i]}\exp(\mathrm{sim}(z_i, z_k)/\tau)} \\
& \mathcal{L}_\mathrm{SimCLR} = \frac{1}{2N_b} \sum_{i=1}^{2N_b} H(\mathbbm{1}_{i'}, p_i)
\end{align*}

where $H$ is the cross-entropy, and $\mathbbm{1}_{i'}$ is the one-hot distribution where all the probability mass is assigned to the index of the positive sample corresponding to $i$, and $\mathrm{sim}$ is the cosine similarity. Intuitively, we are training the model to classify positive
examples in a batch, so the similarity $p$ should be high only
for the true positive.
We introduce the soft objective by replacing the hard positive
distribution $\mathbbm{1}_{i'}$ with a distribution $s_i$. Or, in terms of graphs, we replace the graph from the
\cref{eq:G_ssl} with a soft graph where connection strengths can be any number in $[0,1]$, and, similarly, the distribution $s_i$ and does not have to be one-hot. Considering the example of \cref{fig:idea}, we want the
a photo of a dog to have a representation similar to that of another photo of a dog, somewhat similar to the representation of a cat photo, and different from the representation of a photo of a mug. Given that distribution $s$, we can plug it in directly:

\begin{align*}
& \mathcal{L}_{\mathbb{X}\textnormal{-CLR}} = \frac{1}{2N_b} \sum_{i=1}^{2N_b} H(s_{i}, p_i)
\end{align*}

There are many possible ways to obtain this distribution $s$. 
We could use the meta-data associated with the dataset; in our
case, we utilize a trained text encoder $f_\mathrm{text}$, and encode the text
provided with each image to obtain a representation, which is then
used to calculate similarity between samples $i$ and $j$ using the
cosine similarity. Those pairwise similarities describe the soft graph:

\begin{align*}
\mG^{(\mathrm{soft})}_{i,j} = \mathrm{sim}(f_\mathrm{text}(c_i), f_\mathrm{text}(c_j))
\end{align*}


Where $c_i$ is the caption associated with the $i$-th sample.
The last step before plugging the similarities into the loss function is converting
them to a valid probability distribution using a softmax function:
$$
s_{i,j}=\frac{\exp(\mG^{(\mathrm{soft})}_{i, j}/\tau_s)}{\sum_{k=1}^{2N_b} \exp(\mG^{(\mathrm{soft})}_{i,k}/\tau_s)}
$$

Note that $\tau_s$ is a separate hyperparameter from $\tau$ in the softmax to calculate the learned 
similarities. Higher values of $\tau_s$ put more weight on the 'soft' positives, while lower values in the limit recover the original SimCLR objective.

\section{Experiments}

\subsection{Experimental setup}
We test \xclr{} on three datasets of varying scale: ImageNet \citep{imagenet} (1M), and conceptual captions 3M and 12M \citep{sharma2018conceptual}. We blur faces in all datasets before training our models. We compare to SimCLR \citep{chen2020simple}, to CLIP \citep{radford2021learning} when captions are available. On ImageNet, we compare to SupCon \citep{khosla2020supervised} which uses lables; SCE \citep{denize2023similarity} and ReSSL \citep{zheng2021ressl} which use self-distillation with soft targets, and to SimCLR \citep{chen2020simple}, VICReg \citep{bardes2021vicreg} and BarlowTwins \citep{zbontar2021barlow} which are purely self-supervised algorithms for learning from images. On ImageNet, we also report standard deviations over 5 seeds for the models which we trained in table \cref{tab:imagenet_seeds} (SimCLR, SupCon, and \xclr{}). For the remaining ImageNet models, we took pre-trained encoders.
We use the Sentence Transformer \citep{reimers2019sentence} as the text encoder to construct similarities.
For ImageNet experiments, we generate captions by using the template "a photo of a \_" to generate captions out of class names. In our experiments with the conceptual captions dataset \citep{sharma2018conceptual}, we use the captions as is.
For experiments on ImageNet, we follow SupCon and use
AutoAugment \citep{cubuk2018autoaugment}.  
All experiments on the ImageNet dataset were run for 100 epochs with 
1024 batch size. The learning rate was set to 0.075 for ImageNet models.
For experiments on CC3M and CC12M, we used the standard SimCLR augmentations, and a learning rate of 0.1.
The rest of the settings were kept the same. For more details, see \cref{sec:training_deets}.

In all our experiments, to isolate the effect of our learning objective, we fix the backbone architecture to be a ResNet-50 \citep{he2015deep} model as this is the most widely studied, with optimized hyperparameters, for standard contrastive self-supervised learning \citep{chen2020simple}. We use the same architecture for CLIP's vision encoder and take advantage of already optimized publicly available checkpoints provided by OpenCLIP \citep{openclip} for CC12M. Since no comparable public checkpoint is available for CC3M, we train our own model, see \cref{sec:clip_more}.

\subsection{$\mathbb{X}$-Sample Contrastive with Well-Labeled Samples}

We first experiment with $\mathbb{X}$-Sample Contrastive using well-labeled samples to understand the effect of incorporating similarities across samples in the training objective. 
To do so, we use class labels from ImageNet. 
We compare $\mathbb{X}$-Sample Contrastive ($\mathbb{X}$-CLR) to SimCLR as well as Supervised Contrastive (SupCon), a model whose objective is to explicitly match samples based on their class labels. 
 We evaluate all models across a suite of benchmarks to gauge how well representations generalize in terms of classification performance. 

We find in \cref{tab:imagenet} representations learned via $\mathbb{X}$-CLR improve on standard classification performance, with gains of 12.2\% relative to SimCLR and 1.3\% relative to Supervised Contrastive on ImageNet. We find similar gains when evaluated on revised labels from ImageNet Real of 13.7\% and 1.8\%, respectively. Finally, we find by capturing similarities across samples, representations learned via $\mathbb{X}$-CLR are more capable of disambiguating objects from backgrounds and attributes with gains on ImageNet-9 (for details see \cref{sec:inet9}) \citep{xiao2020noise} and ObjectNet \citep{barbu2019objectnet}.
 
\begin{table}[]
\centering
\caption{\textbf{$\mathbb{X}$-Sample Contrastive loss outperforms contrastive (SimCLR) and even Supervised Contrastive with ImageNet pretraining.}}
\label{tab:imagenet}
\resizebox{\columnwidth}{!}{%
\begin{tabular}{lccccccc}
\toprule
               & \multicolumn{1}{l}{} & \multicolumn{1}{l}{} & \multicolumn{2}{l}{Background Decomposition} & \multicolumn{1}{l}{} & \multicolumn{2}{c}{MIT States} \\ \cmidrule(lr){4-5} \cmidrule(lr){7-8} 
Method         & ImageNet & ImageNet Real & Same Class & Mixed & ObjectNet & Objects & Attributes \\ \midrule
SCE            & 71.3     & 78.7         & 61.7       & 58.4       & 20.2      & 44.5                 & 31.0   \\
ReSSL (1 crop) & 69.4     & 76.9         & 56.3       & 53.2       & 18.3      & 44.5                 & 31.2   \\
VICReg         & 72.4     & 79.0         & 60.8       & 56.8       & 20.5      & 43.5                 & 26.9   \\
Barlow Twins   & 72.8     & 80.0         & 62.7       & 59.4       & 21.6      & \textbf{45.9}        & \textbf{31.7}   \\
SimCLR & 63.4 & 67.8  & 44.7 & 38.9 &  12.1 & 40.9  & 29.1 \\          
SupCon & 74.3 & 79.7  & 64.0 & 59.9 & 24.4 & 45.6  & 30.8 \\       
\xclr{} & \textbf{75.6} & \textbf{81.5} & \textbf{66.6} & \textbf{62.7} & \textbf{27.5} & \textbf{45.9} & 31.1  \\
\bottomrule
\end{tabular}%
}
\end{table}


\begin{figure}
\centering
\begin{subfigure}[t]{0.32\linewidth}
\includegraphics[width=\textwidth]{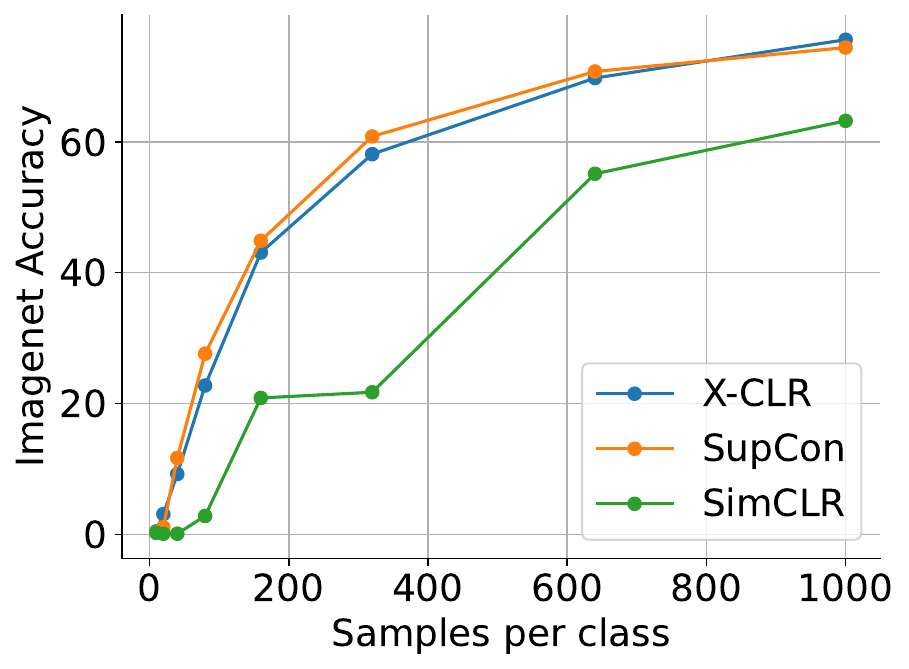}   
\caption{}
\label{fig:imagenet-data-efficiency}
\end{subfigure}
\hfill
\begin{subfigure}[t]{0.32\linewidth}
\includegraphics[width=\textwidth]{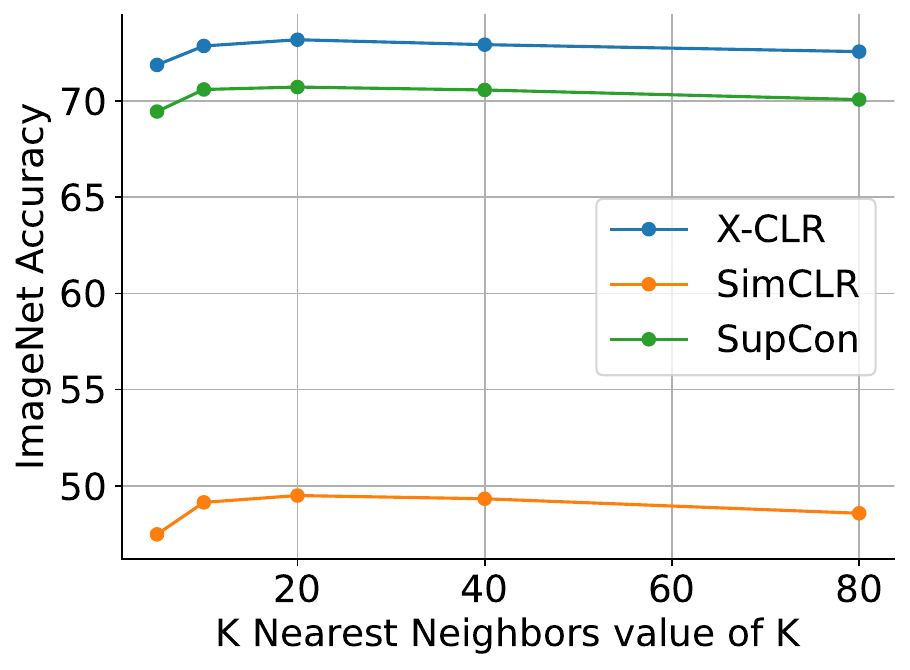}   
\caption{}
\label{fig:knn-imagenet}
\end{subfigure}
\hfill
\begin{subfigure}[t]{0.32\linewidth}
\centering
\includegraphics[width=\textwidth]{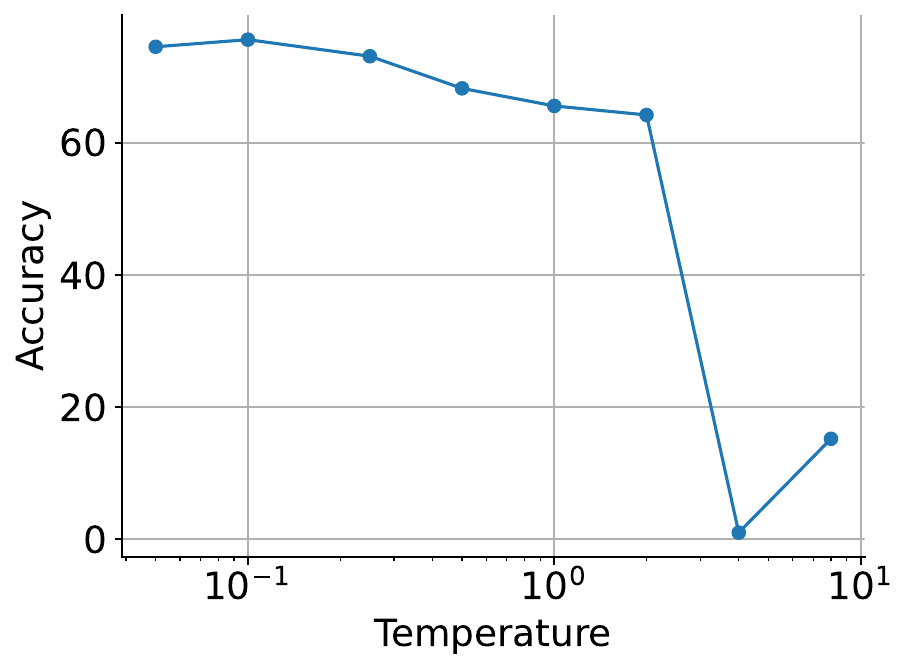}
\caption{}
\label{fig:sensitvity}
\end{subfigure}
\caption{\textbf{(a) $\mathbb{X}$-Sample Contrastive Loss is data efficient with ImageNet pretraining.} We outperform SimCLR in low data regimes and match Supervised Contrastive trained on ground truth labels at varying levels of data scarcity. \textbf{(b) KNN performance ImageNet.} $\mathbb{X}$-CLR outperforms other methods with KNN probing for a range of values of K. \textbf{(c) Sensitivity of $\mathbb{X}$-Sample Contrastive to temperature.} We test the performance of our method when trained with different values of temperature $\tau_s$ on ImageNet data.}
\end{figure}

\paragraph{Can we improve contrastive learning under data scarcity?} To answer this question, we train all three models SimCLR, SupCon, and $\mathbb{X}$-CLR by varying the number of samples seen for each class in ImageNet. We find $\mathbb{X}$-CLR, by incorporating information about class labels and how they relate, is able to learn representations that improve over the performance of SupCon trained with ground truth class labels and outperform SimCLR even when few training samples are available per class as shown in \cref{fig:imagenet-data-efficiency}.

\subsection{$\mathbb{X}$-Sample Contrastive with Noisy Multimodal Samples}

Contrastive loss also plays a pivotal role in multimodal vision-language models such as CLIP. The contrastive training objective matches noisy caption-image pairs. Here we experiment with $\mathbb{X}$-Sample Contrastive by using the noisy captions to learn similarities across samples. We compare both SimCLR as a standard contrastive model and CLIP trained on the same caption-image data across two levels of scale: 3 and 12 million samples from CC3M and CC12M. 

We find incorporating $\mathbb{X}$-Contrastive leads to representations with higher classification accuracy and disambiguation of objects from their attributes and backgrounds. With CC12M training  shown in \cref{tab:cc12m},  $\mathbb{X}$-Contrastive outperforms SimCLR by 0.5\% and CLIP by 0.6\% with CC12M with similar gains for ImageNet Real. We also find $\mathbb{X}$-CLR training can better disambiguate object foreground from backgrounds, with gains of 0.6-1.5\% over SimCLR and 3.3-5.6\% over CLIP.

We find learning similarites across samples with $\mathbb{X}$-CLR leads to more considerable gains when less data is available. $\mathbb{X}$-CLR outperforms SimCLR by 1.2\% and CLIP by 17.2\% on ImageNet, with similar gains on ImageNet Real as shown in \cref{tab:cc3m}. We find $\mathbb{X}$-CLR training can more considerably disambiguate object foregrounds from backgrounds compared to CLIP when less training data is available, with gains of 10.3-14.2\% over CLIP.

\begin{table}[]
\caption{\textbf{$\mathbb{X}$-Sample Contrastive with CC3M training outperforms contrastive baselines.}
}
\label{tab:cc3m}
\centering
\begin{tabular}{lccccc}
\toprule
                &               &               & \multicolumn{2}{l}{Background Decomposition} &               \\ \cmidrule(lr){4-5}
Method          & ImageNet      & ImageNet Real & Same Class            & Mixed                & ObjectNet     \\ \midrule
SimCLR & 57.0            & 64.0            & 24.4          & 18.9          & 10.8          \\
CLIP & 41.0  & 47.6  & 12.5 & 10.6 & 7.8  \\
\xclr{} & \textbf{58.2} & \textbf{65.6} & \textbf{26.7} & \textbf{20.3} & \textbf{11.5} \\
\bottomrule
\end{tabular}%

\end{table}

\begin{table}[]
\centering
\caption{\textbf{$\mathbb{X}$-Sample Contrastive with CC12M training outperforms contrastive baselines.}}
\label{tab:cc12m}
\begin{tabular}{lccccc}
\toprule
\multicolumn{1}{l}{} &             &               & \multicolumn{2}{c}{Background Decomposition} &               \\ \cmidrule(lr){4-5}
Method               & ImageNet    & ImageNet Real & Same Class            & Mixed                & ObjectNet     \\
\midrule
SimCLR                   & 58.9                 & 66                   & 24.6                 & 19.8                 & 12.7           \\
CLIP                     & 58.8        & 66.1        & 20.5        & 17.1        & 11.9  \\
\xclr{} & \textbf{59.4}        & \textbf{66.7}        & \textbf{26.1}        & \textbf{20.4}        & \textbf{13.4}    \\
\bottomrule
\end{tabular}%
\end{table}

\subsection{$\mathbb{X}$-Sample Contrastive can be used to finetune pretrained backbones}

\begin{table}[]
\centering
\caption{\textbf{$\mathbb{X}$-CLR can be used to finetune pretrained models.}}
\label{tab:finetuning-imagenet}
\resizebox{\columnwidth}{!}{%
\begin{tabular}{lccccc}
\toprule
\multicolumn{1}{l}{} &             &               & \multicolumn{2}{c}{Background Decomposition} &               \\ \cmidrule(lr){4-5}
& ImageNet      & ImageNet Real  & Same Class & Mixed & ObjectNet      \\
\midrule
SimCLR & 63.4           & 67.8          & 44.7          & 38.9        &  12.1 \\          
+ \xclr{} finetuning & \textbf{66.5} & \textbf{74.4} & \textbf{53.9} & \textbf{50.0} & \textbf{17.4} \\ 
\bottomrule
\end{tabular}%
}
\end{table}

We validate whether $\mathbb{X}$-CLR can be used as a finetuning objective for pretrained backbones, given the growing abundance of publicly available backbones. Here, we evaluate a pretrained SimCLR model by finetuning for 10 epochs on ImageNet with $\mathbb{X}$-CLR instead of the original SimCLR contrastive objective. We see in \cref{tab:finetuning-imagenet} finetuning with $\mathbb{X}$-CLR improves classification performance on ImageNet by 3.1\% and on ImageNet Real by 6.6\%. Furthermore, we see by relating samples during the finetuning stage, $\mathbb{X}$-CLR can disambiguate object foregrounds from backgrounds with grains of 9.2-11.1\% on ImageNet-9 as well as improvements on natural object transformations from ObjectNet with a gain of 5.3\% after finetuning.

\subsection{$\mathbb{X}$-Sample Contrastive introduces only minimal computational overhead}
Both for ImageNet and conceptual captions datasets, we don't run the text encoder
for each sample we see, and instead precompute the similarity values. For more details, see \cref{sec:training_deets}. Avoiding running the text encoder during model training avoids the extra overhead at the price of some pre-processing. Pre-processing takes less than 2 hours for CC12M when using one GPU, about 30 minutes for CC3M, and less than 5 minutes for ImageNet. To further analyze how much overhead there is, we compare the average
time it takes to process one batch for SimCLR and $\mathbb{X}$-CLR. The results are shown in \cref{tab:time-efficiency}.
Overall, we didn't notice any significant difference in the amount of time it takes to train models with the $\mathbb{X}$-CLR objective compared to the regular contrastive objective. To train on ImageNet, we used 8 Nvidia V100s, and each run took about 30 hours. With the same setup, CC3M runs took about 50 hours, and CC12M runs took roughly 9 days. 

\begin{table}[]
\centering
\caption{\textbf{Analyzing the computation overhead of the $\mathbb{X}$-Sample Contrastive objective during training.} $\mathbb{X}$-CLR introduces nearly no computational overhead compared to SimCLR. }
\label{tab:time-efficiency}
\begin{tabular}{lllll}
\toprule
Method         & Seconds per batch ImageNet & Seconds per batch CC &  &  \\
\midrule
SimCLR         & 0.866 ± 0.008              & 0.874 ± 0.034        &  &  \\
$\mathbb{X}$-CLR & 0.866 ± 0.010              & 0.877 ± 0.032        &  &  \\
\bottomrule
\end{tabular}
\end{table}

\section{Analyzing representations learned via $\mathbb{X}$-Sample Contrastive}

\subsection{KNN Clustering}

To confirm the representations learned via $\mathbb{X}$-CLR also work well for downstream tasks with non-linear decision boundaries, we perform evaluation using the common K-nearest neighbor (KNN) protocol.
The results shown in \cref{fig:knn-imagenet} demonstrate $\mathbb{X}$-CLR outperforms both SimCLR and SupCon baselines across a range of choices for $K$. We also show KNN results for models trained on conceptual captions in \cref{sec:knn-eval-ap}.

\subsection{Visualizing the learned graph from $\mathbb{X}$-Sample Contrastive representations}

Here we examine whether the learned representations from $\mathbb{X}$-Sample Contrastive capture semantically meaningful similarities.
To do so, we select four groups of three ImageNet classes: felines, dogs, types of balls, and musical instruments.
For each pair of classes, we then compare the representation similarities using cosine similarity. A higher average pairwise similarity indicates the model's latent representations encode the classes similarly.
In \cref{fig:learned-graph} we show the graph of similarities learned after training with $\mathbb{X}$-CLR on ImageNet. 
We find that the image encoder successfully captures the similarity within the class groups.

\subsection{The effect of softmax temperature, and inferred similarity graph}

We also examine the effect of hyperparameter choices. We show the sensitivity of \xclr{} to temperature $\tau_s$ in \cref{fig:sensitvity} on ImageNet. In the limit, when temperature goes to 0, we recover Supervised Contrastive method for ImageNet, or SimCLR in case of conceptual captions. With low temperature, the similarity is 1 only if the captions are exactly the same. As 
the temperature increases, more weight is put on the soft positives compared to the true positives (i.e. augmentations
of the same sample). With high temperature, our method is unstable as too much emphasis is put on the soft positive examples compared to the true positives. We find that the value of 0.1 strikes the optimal balance and provides an improvement over pure Supervised Contrastive objective, while still emphasizing true positives enough. For more details 
regarding how $\tau_s$ changes the objective, see \cref{fig:diag_mass}.

We also experiment with different ways of inferring the graph, including using different text encoders,
using WordNet \citep{wordnet} hierarchy distance, and the purely random graph. We find that overall, calculating the similarities using the sentence transformer worked the best \citep{reimers2019sentence}. A more detailed comparison of different graph sources can be found in \cref{sec:different-similarities}.

\subsection{The impact of label quality for fine-grained attribute disambiguation}

we show in \cref{tab:label-quality-mit-states} how label quality can impact downstream performance on finer-grained attribute disambiguation. we find larger labels from noisy captions degrades performance for fine-grained object attributes in mit states \citep{isola2015discovering} for both contrastive and clip. we find \xclr{} with high quality labels from imagenet, can outperform models trained on much larger noisier data. compared to clip trained on 12$\times$ larger data, \xclr{} achieves 30.9\% vs. 23.3\% for clip on attribute classification and 45.8\% vs. 36.9\% for clip on object classification under different states. to see more details regarding the mit states evaluation, see \cref{sec:training_deets}.

\begin{table}[]
\centering
\caption{\textbf{Label quality matters for fine-grained attribute disambiguation.} \Vlad{change to blurred results}}
\label{tab:label-quality-mit-states}
\resizebox{\columnwidth}{!}{%
\begin{tabular}{lcccc}
\toprule
Pretraining             & Data Size & Quality & MIT States Attributes & MIT States Objects \\
\midrule
CLIP CC3M                 & 3M        & Noisy   & 27.0              & 40.1       \\
CLIP CC12M                & 12M       & Noisy   & 23.3              & 36.9       \\
$\mathbb{X}$-CLR CC3M     & 3M        & Noisy   & 29.5             	& 40.7       \\
$\mathbb{X}$-CLR CC12M    & 12M       & Noisy   & 30.1             	& 42.1       \\
$\mathbb{X}$-CLR ImageNet & 1M        & High    & \textbf{30.9}    	& \textbf{45.8}       \\

\bottomrule
\end{tabular}%
}
\end{table}

\begin{figure}
\centering
\includegraphics[width=\textwidth]{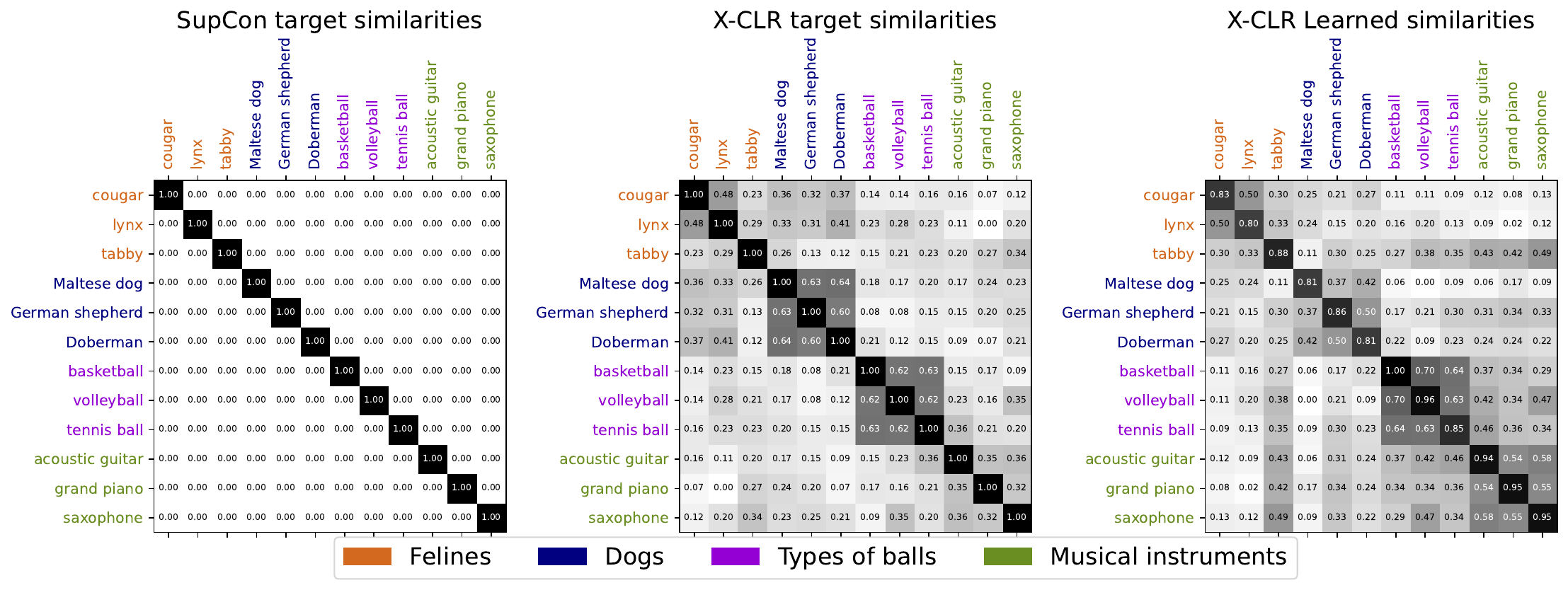}
\caption{\textbf{Visualizing pairwise similarities} SupCon \citep{khosla2020supervised} 
objective does not encourage non-zero similarity between samples of different classes (left), while $\mathbb{X}$-CLR target similarities take into account semantic closeness within categories such as dogs or types of balls (center). On the right, we see that the trained model successfully learns the soft similarity. For more graphs, see \cref{fig:sims_all}.}
\label{fig:learned-graph}
\end{figure}

\section{Discussion}

In the present work, we have proposed a new graph perspective on the commonly used contrastive learning methods and used our insights to develop a better learning objective, $\mathbb{X}$-CLR, by using a soft similarity graph.
The adjacency matrix of the proposed graph contains not just 0 and 1, but also any values between,
with the ability to capture the degree of similarity \textit{across} samples. We experiment with different ways of constructing the
graph, and find that indeed we can build a soft graph that improves over the existing binary graph contrastive methods. However, we believe that there are better ways of constructing the graph than
what we found, particularly for the conceptual captions dataset where the captions are quite noisy. 
A better graph can possibly be built using other metadata, such as location or time. We also
believe that ideas from $\mathbb{X}$-CLR can possibly be integrated into non-contrastive objectives such as BYOL \citep{grill2020bootstrap} or VICReg \citep{bardes2021vicreg} to enrich representations with similarities across samples.

\textbf{Limitations} \, The main limitation of the present work is that constructing the cross-sample similarity graph requires extra data, as well as some extra memory to store it. When the extra data is not available, the only options remaining are to build the graph using the augmentations, 
self-distillation, or other pre-trained models. The resulting method is also highly dependent on the quality of the graph, as we have seen with conceptual captions datasets.

\section{Acknowledgements}
This material is based upon work supported by the National Science Foundation under NSF Award 1922658.
\clearpage

\bibliography{main}
\bibliographystyle{plainnat}

\newpage

\appendix

\section{Appendix / supplemental material}

\begin{table}[]
\centering
\caption{\textbf{Analyzing statistical significance of ImageNet results.} Each experiment is ran with 5 seeds, we report the mean and standard deviation.}
\label{tab:imagenet_seeds}
\resizebox{\columnwidth}{!}{%
\begin{tabular}{lccccccc}
\toprule
               & \multicolumn{1}{l}{} & \multicolumn{1}{l}{} & \multicolumn{2}{l}{Background Decomposition} & \multicolumn{1}{l}{} & \multicolumn{2}{c}{MIT States} \\ \cmidrule(lr){4-5} \cmidrule(lr){7-8} 
Method         & ImageNet & ImageNet Real & Same Class & Mixed & ObjectNet & Objects & Attributes \\ \midrule
SimCLR & 63.43 ± 0.12 & 67.75 ± 0.27   & 12.07 ± 0.33 & 38.88 ± 0.43 & 44.67 ± 0.60 & 40.92 ± 0.26           & 29.08 ± 0.17  \\          
SupCon & 74.30 ± 0.16 & 79.66 ± 0.12   & 24.42 ± 0.25 & 59.08 ± 0.44 & 64.00 ± 0.62 & 45.56 ± 0.16           & 30.83 ± 0.20   \\       
\xclr{} & 75.56 ± 0.09 & 81.54 ± 0.13  & 27.53 ± 0.13 & 62.74 ± 0.27 & 66.59 ± 0.25 & 45.86 ± 0.15  & 31.10 ± 0.18   \\
\bottomrule
\end{tabular}%
}
\end{table}

\subsection{More learned similarities comparisons}
\label{sec:different-similarities}
We compare inferring the similarity graph using different text encoders: 
\begin{itemize}
    \item Graph with connections only between samples of the same class (SupCon);
    \item Graph with connections only between augmentations of the same image (SimCLR);
    \item Graph where soft similarity is inferred by comparing representations of the sample captions. The representations are computed using the sentence transformer \citep{reimers2019sentence}, CLIP text encoder \citep{radford2021learning}, LLama2 encoder \citep{touvron2023llama};
    \item Graph where the connection strength is defined by the distance in WordNet \citep{wordnet} hierarchy;
    \item Random graph where the cross-sample connections' strengths are fully random;
\end{itemize}

\begin{figure}
\centering
\begin{subfigure}[t]{0.8\linewidth}
    \includegraphics[width=\textwidth]{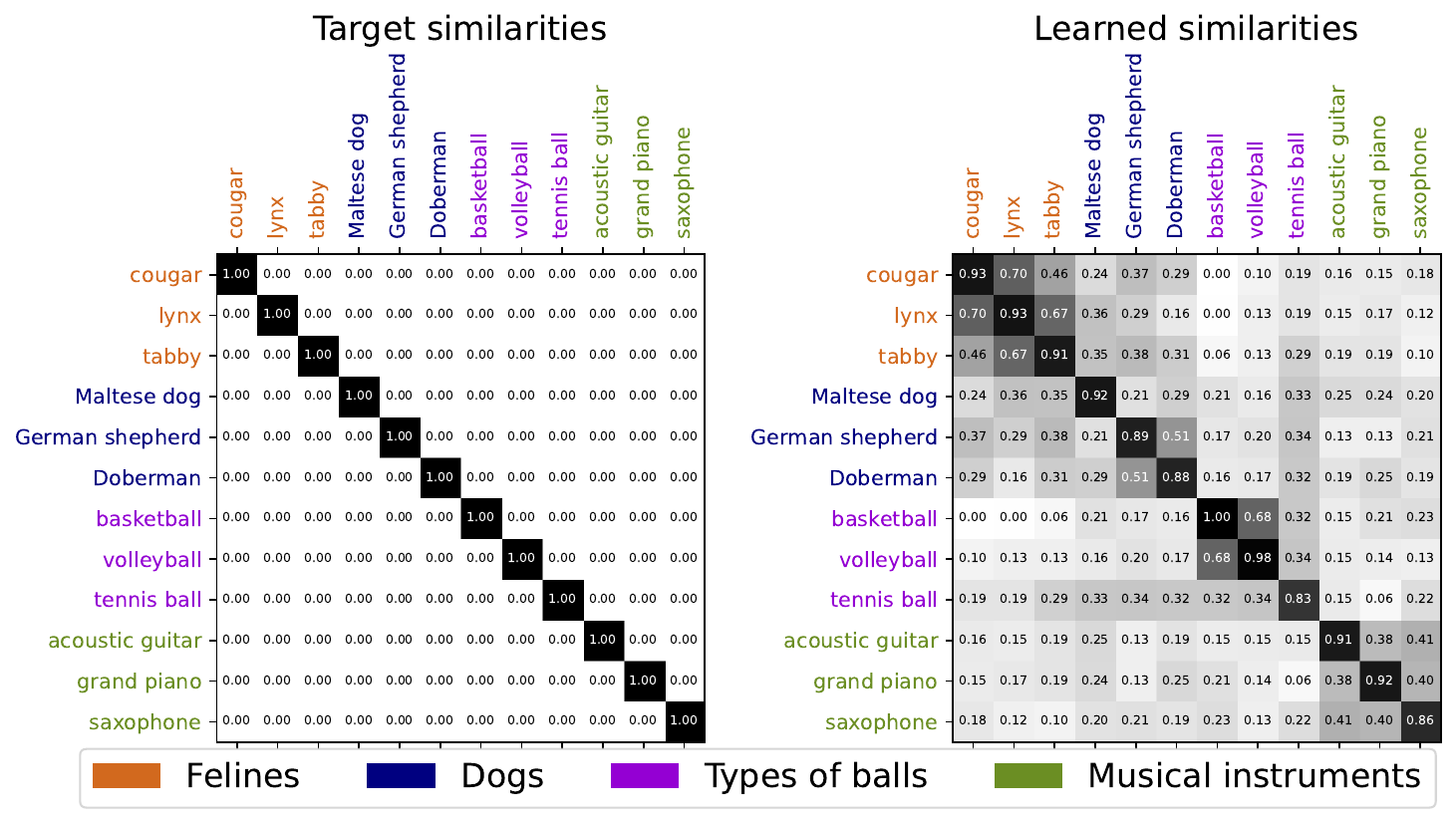}
    \caption{SupCon target and learned similarities}
    \label{fig:sims_supcon}
\end{subfigure}
\begin{subfigure}[t]{0.8\linewidth}
    \includegraphics[width=\textwidth]{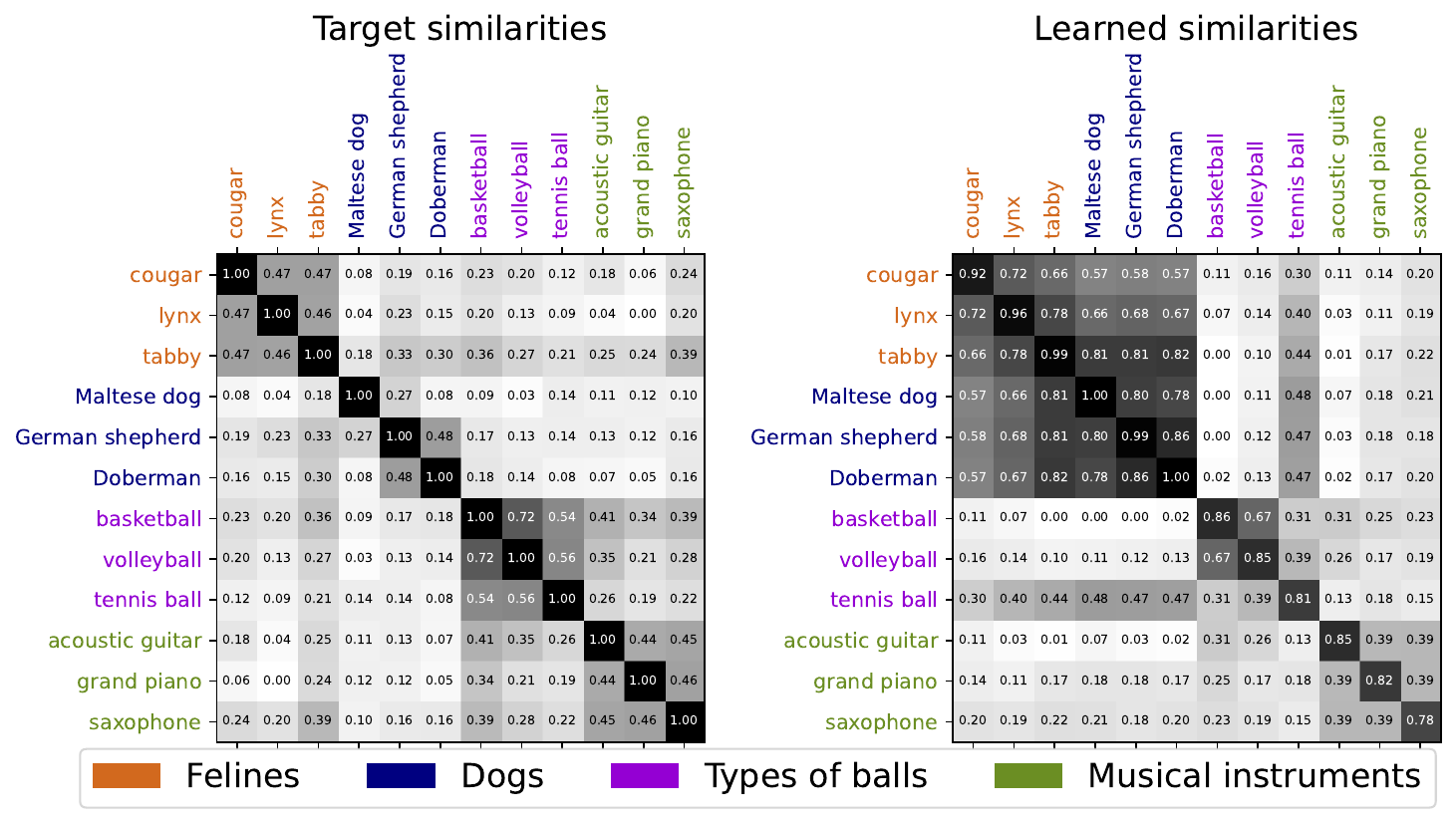}
    \caption{CLIP target and learned similarities}
    \label{fig:sims_clip}
\end{subfigure}
\caption{Target and learned similarities for different graphs.}
\label{fig:sims_all}
\end{figure}

The results are shown in \cref{tab:similarities}. We find that overall, the Sentence Transformer graph performs the best,
although the CLIP text encoder achieves good performance as well. Interestingly, we find that using WordNet hierarchy distance did not work well.
We visualize learned and target similarities for SupCon graph and for the graph built using CLIP text encoder in \cref{fig:sims_all}.

\paragraph{Visualising similarities} In \cref{fig:learned-graph}, to visualize learned similarities, for 
each class we pick 100 examples from the dataset, encode them. Then, to calculate the average learned similarity between
two classes, we take the 100 examples for each of the two classes, and calculate the Cartesian product, yielding 10,000 similarities. We take the mean over those 10,000 similarities to represent the average learn similarity for a class pair.

\begin{figure}
\centering
\begin{subfigure}[t]{0.41\linewidth}
\includegraphics[width=\textwidth]{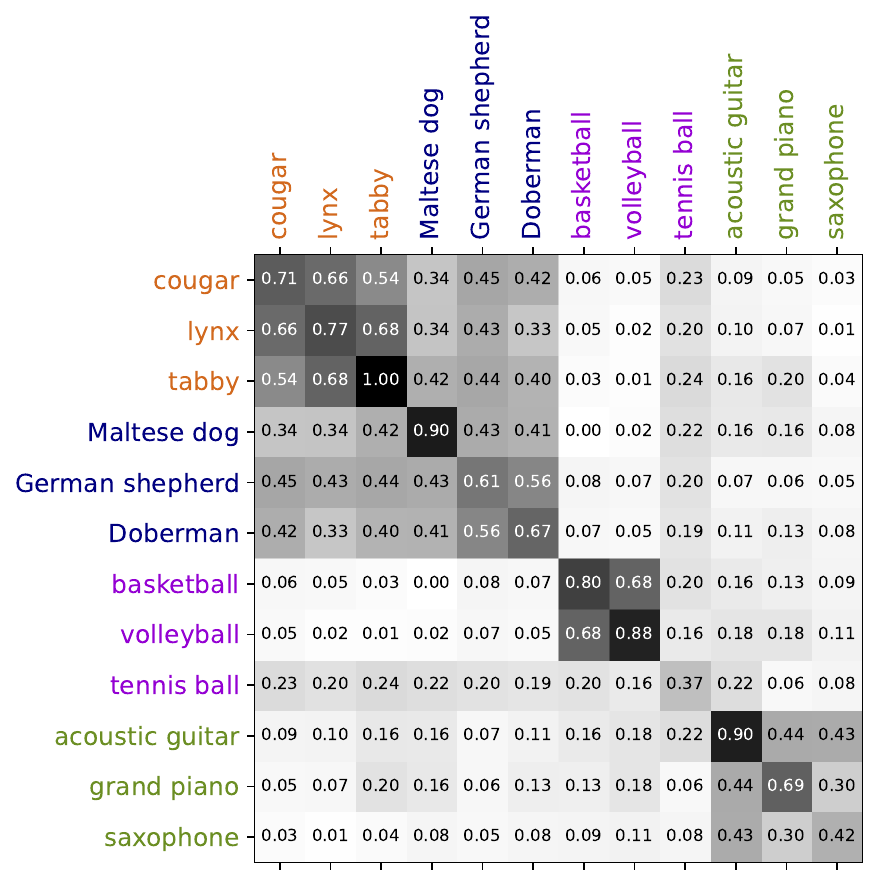}
\caption{CC3M similarities}
\end{subfigure}
\begin{subfigure}[t]{0.41\linewidth}
\includegraphics[width=\textwidth]{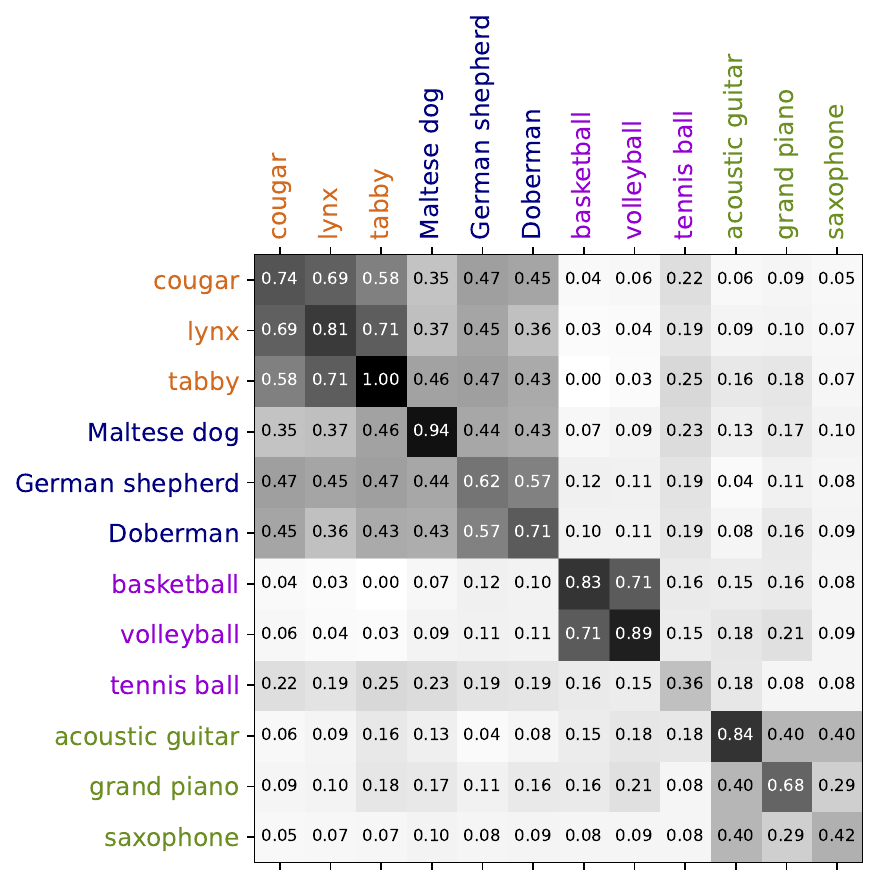}
\caption{CC12M similarities}
\end{subfigure}
\label{fig:sim_cc}
\caption{\xclr{} Learned similarities when trained on \textbf{a)} CC3M and \textbf{b)} CC12M.}
\end{figure}

\paragraph{Similarities when training on CC datasets} In \cref{fig:sim_cc}, we show the similarities learned by \xclr{} on CC3M and CC12M datasets.

\subsection{Analyzing statistical significance of the results}
To make sure the difference in performance we observe is statistically significant, we run \xclr{}, SimCLR, and SupCon 
pretraining with 5 different seeds. We report the results of the evaluations in \cref{tab:imagenet_seeds}. 

\begin{table}[]
\centering
\caption{The effect of the similarity source on the model performance.}
\label{tab:similarities}
\resizebox{\columnwidth}{!}{%
\begin{tabular}{lrrrrr}
\toprule
               & \multicolumn{1}{l}{} & \multicolumn{1}{l}{} & \multicolumn{2}{l}{Background Decomposition} & \multicolumn{1}{l}{} \\ \cmidrule(lr){4-5}
Similarity source             & ImageNet & ImageNet Real & Same Class & Mixed & ObjectNet \\
\midrule
Augmentation graph (SimCLR)             & 63.2     & 67.5          & 45.5       & 38.3  & 12.5       \\
Sentence Transformer (\xclr{})               & 75.6     & 81.6          & 66.5       & 62.3  & 27.7       \\
CLIP text   encoder                          & 74.4     & 80.6          & 67.5       & 64.2  & 24.5       \\
LLama2 text encoder                          & 40.9     & 45.8          & 38.3       & 36.0  & 4.3        \\
Random per class pair                        & 74.5     & 80.8          & 71.0       & 68.0  & 26.6       \\
Random per sample pair                       & 0.1      & 0.1           & 0          & 0     & 0          \\
True class graph (SupCon)                    & 74.4     & 79.7          & 63.3       & 58.8  & 24.1       \\
Distance in WordNet hierarchy                & 68.3     & 74.9          & 55.7       & 52.1  & 21.2       \\
\bottomrule
\end{tabular}
}
\end{table}

\subsection{Analyzing the learned graph}

\begin{table}[]
\centering
\caption{Analyzing the learned representations' connectivity}
\label{tab:gen_analysis}
\begin{tabular}{lrrrrr}
\toprule
Metric                        & CLIP  & SimCLR & SupCon & X-CLR \\
\midrule
Label error (↓)               & 0.550 & 0.250  & 0.250  & 0.223 \\
Intra-class connectivity  (↑) & 1.233 & 1.700  & 2.005  & 2.193 \\
\bottomrule
\end{tabular}
\end{table}

We follow the analysis of \citet{zhang2023generalization} and results in \cref{tab:gen_analysis}. The analysis studies two values: label error which measures how similar samples of different classes are on average, and intra-class connectivity, which measures the similarity of the samples within the class relative to those from different classes. This allows us to determine how well the learned graph captures the class relationships in the data.
Since the open-source repository of that paper did not contain the code for analysis, we re-implemented it to the best of our ability.

According to these metrics, \xclr{} representation is the best among the baselines.
We note that our SimCLR numbers are much better than in the original paper. We suspect that it’s due to the fact that the authors train SimCLR with the batch size of 512, while we use 2048. ImageNet classification performance of our SimCLR model is also higher, at 63.4, compared to 61.2.

We also note that label error, which is the measure of average similarity between instances of different classes, is lower for our method, although the loss itself encourages it to be higher for related samples. This is due to the fact that in this analysis, we use the first 10 classes from ImageNet (replicating the original procedure), and those classes are not related to each other.

\subsection{KNN evaluation}
\label{sec:knn-eval-ap}
\begin{figure}
\centering
\includegraphics[width=\textwidth]{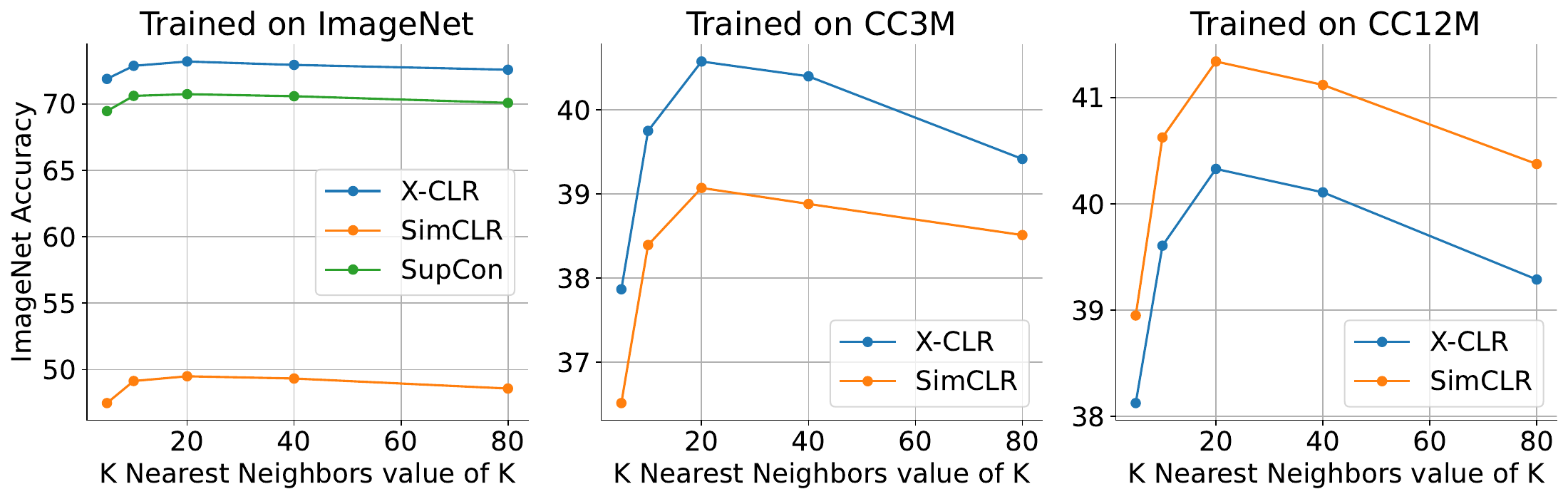}
\caption{Results of models trained on ImageNet, CC3M, CC12M on ImageNet validation when using KNN classifier.}
\label{fig:knn_all_imagenet}
\end{figure}

Apart from testing the models trained on ImageNet using KNN, we also evaluate the models trained on CC3M and CC12M. The results are shown in \cref{fig:knn_all_imagenet}.
We see that \xclr{} performs better on CC3M, and comparatively with SimCLR when trained on CC12M.

\subsection{ImageNet-9 details}
\label{sec:inet9}
ImageNet-9 \citep{xiao2020noise} proposes multiple benchmarks to test model robustness to the background perturbation.
In our work, we use "Mixed-Same" and "Mixed-Rand" tasks from ImageNet-9, and refer to them together as "Background Decomposition".

\subsection{CLIP details}
\label{sec:clip_more}
In CC3M experiments, we train the model from scratch, as OpenCLIP didn't have a checkpoint trained on that dataset.
We trained both for 32 and 100 epochs, and found that the model trained for 100 epochs performs better.
Since 32 epochs is the default CLIP number of epochs, we also report results for 32 epochs. The results are
shown in \cref{tab:clip32vs100}.

\begin{table}[]
\centering
\caption{\textbf{CLIP on CC3M} We train our own models on CC3M and find that training longer improves the
performance. Nevertheless, CLIP struggles with small datasets.}
\label{tab:clip32vs100}
\begin{tabular}{lccccc}
\toprule
                &               &               & \multicolumn{2}{l}{Background Decomposition} &               \\ \cmidrule(lr){4-5}
Method          & ImageNet      & ImageNet Real & Same Class            & Mixed                & ObjectNet     \\ \midrule
CLIP 100 epochs & 41.0  & 47.6  & 12.5 & 10.6 & 7.8  \\
CLIP 32 epochs  & 36.8  & 42.0  & 11.5 & 9.8  & 6.0 \\
\bottomrule
\end{tabular}%

\end{table}

\subsection{More training details}
\label{sec:training_deets}
We train SimCLR, SupCon and \xclr{} using the LARS optimizer \citep{you2017large}.
In all cases, we use the same ResNet-50, with a two layer projector on top. The output dimension of the projector is 128.

\paragraph{Fetching similarities} For ImageNet, since the number of classes is known, we pre-compute the similarity matrix of dimension
$1000 \times 1000$, and retrieve elements from it depending on the associated class
labels for a given sample pair to obtain the similarity value.
For conceptual captions, we run the text encoder on the full dataset and save the encodings to disk. Then, when loading an image from disk, we also load the associated
encoding of the corresponding caption. The similarity matrix for a given batch is then
obtained by calculating the Cartesian product of those encodings.

\paragraph{MIT States} In order to evaluate on this dataset using linear probing, we split the dataset randomly into two
even parts, one used for training the linear layer, the other for evaluation. We train separately to classify objects and
attributes.

\subsection{Understanding similarities}

\begin{figure}
\centering
\begin{subfigure}[t]{0.6\linewidth}
\includegraphics[width=\textwidth]{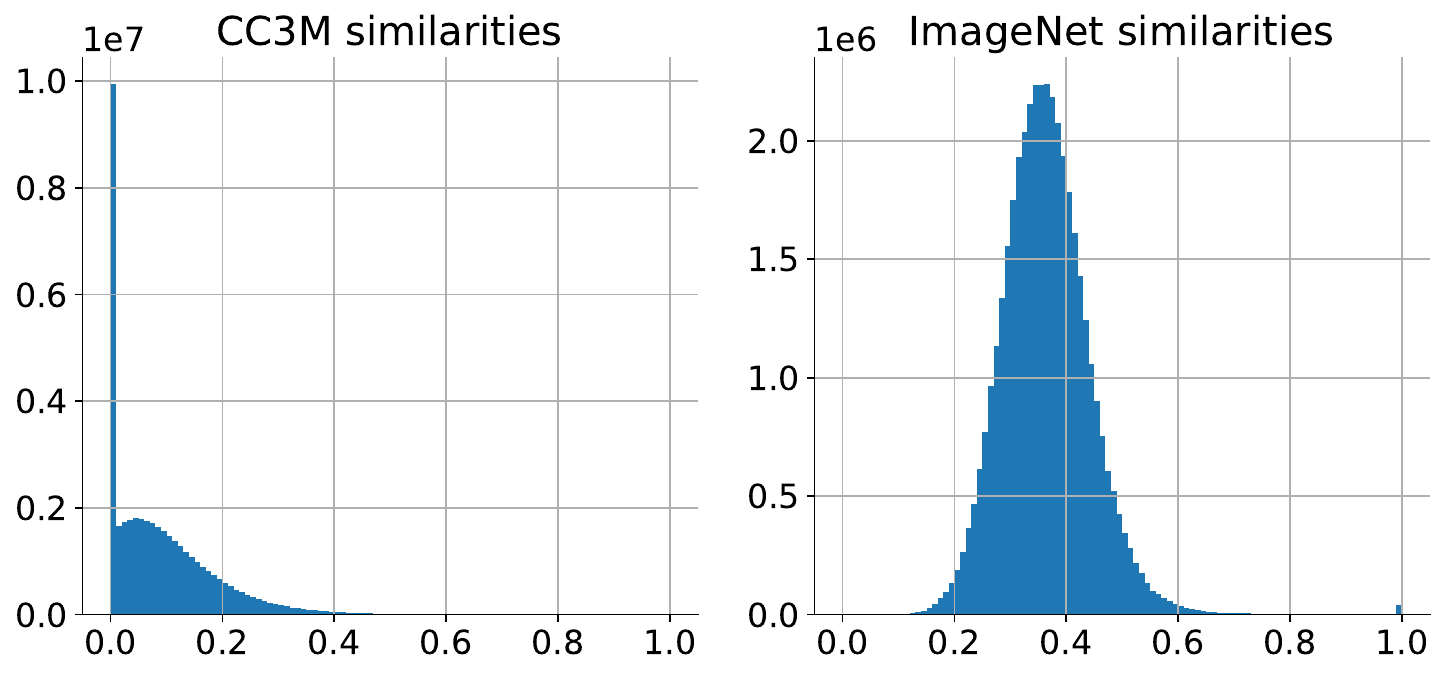}
\caption{}
\label{fig:similarity_histogram}
\end{subfigure}
\begin{subfigure}[t]{0.38\linewidth}
\includegraphics[width=\textwidth]{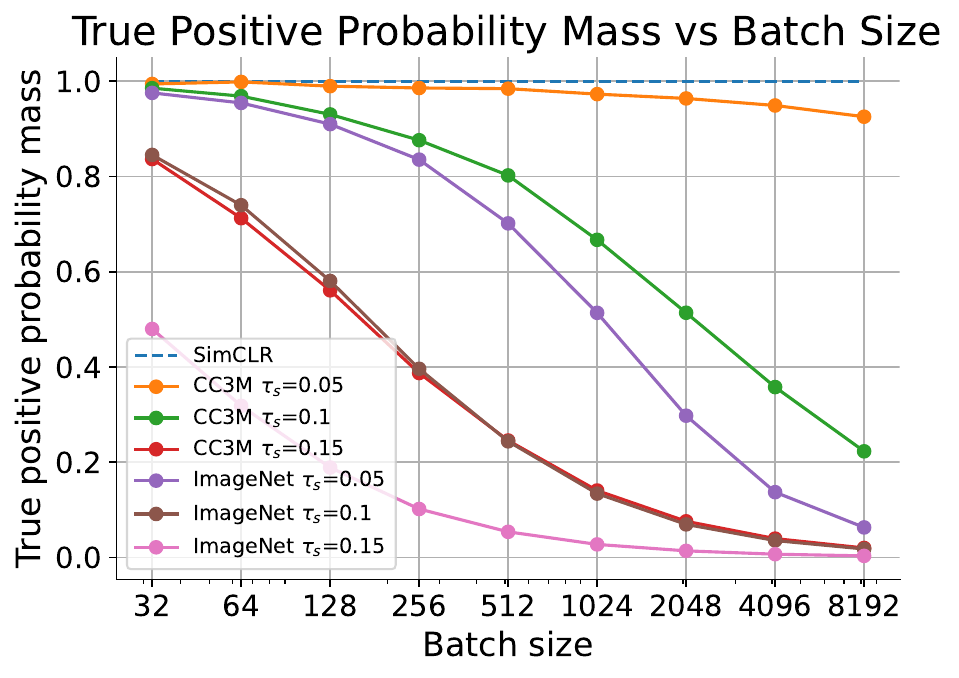}
\caption{}
\label{fig:diag_mass}
\end{subfigure}
\caption{(a) Histograms of the similarities calculated using Sentence Transformer on ImageNet and CC3M. While for 
ImageNet the average similarity is around 0.35, it is much lower on CC3M, signifying that the graph contains
less information for CC3M. (b) Effect of the temperature and batch size on the weight assigned to the true positvie.}
\end{figure}

To understand the graphs we built using for different datasets, we investigate the average cross-sample 
similarity in the dataset. The result is shown in \cref{fig:similarity_histogram}. We find that CC3M similarities are in
general lower, possibly because of lower quality annotations. We also investigate how much weight is assigned to the
true positive examples. For SimCLR, it's always 1. For our method, the amount of similarity assigned to other samples in the
batch depends on the temperature $\tau_s$, and the batch size. The exact relationship is shown in \cref{fig:diag_mass}.

\subsection{Connection between Supervised Contrastive Learning and \xclr{}}

Here, we will outline how as the temperature $\tau_s$ approaches 0, \xclr{} becomes SupCon.
Supervised Contrastive Learning \citep{khosla2020supervised} also uses image augmentations, and augments each image twice, to obtain what they call "a multiviewed batch". Then, in equation 2, they propose the loss:
$$
\mathcal{L}_\mathrm{out}^\mathrm{sup}=\sum_{i\in I } \mathcal{L}_{\mathrm{out},i}^\mathrm{sup} = \sum_{i\in I}\frac{-1}{| P(i) |} \sum_{p\in P(i)} \log p_{i,p}
$$
where $p_{i,j}$ is defined as follows:
$$p_{i,j}=\frac{\exp(\mathrm{sim}(z_i, z_j)/\tau)}{\sum_{i=1}^{2N_b} \mathbb{1}_{[k \neq i]}\exp(\mathrm{sim}(z_i, z_k)/\tau)}$$
However, $| P(i) |$ is exactly the number of positive samples, and $p_{i, p}$ is the probability of $i$ and $p$ being a positive pair according to the model. We set $s_i^\mathrm{supcon}$ to be a distribution over $2N_b - 1$ candidates for positive pairs and define it as follows:
$$
s_{i,j}^\mathrm{supcon}= \begin{cases} \frac{1}{|P(i)|}, & \text{if } j \in P(i) \\ 0, & \text{otherwise} \end{cases}
$$
Then, we can write down the original loss as:
$$
\mathcal{L}_{\mathrm{out},i}^\mathrm{sup} = H(s_i^\mathrm{supcon}, p_i)
$$
where H is the cross-entropy. This looks exactly like the \xclr{} objective. We can recover SupCon objective if we increase the temperature $\tau_s$: the resulting distribution $s_i$ will be equal to $s_i^\mathrm{supcon}$.

\subsection{T-SNE of the learned representations}

\begin{figure}
\begin{subfigure}[t]{0.31\linewidth}
\includegraphics[width=\textwidth]{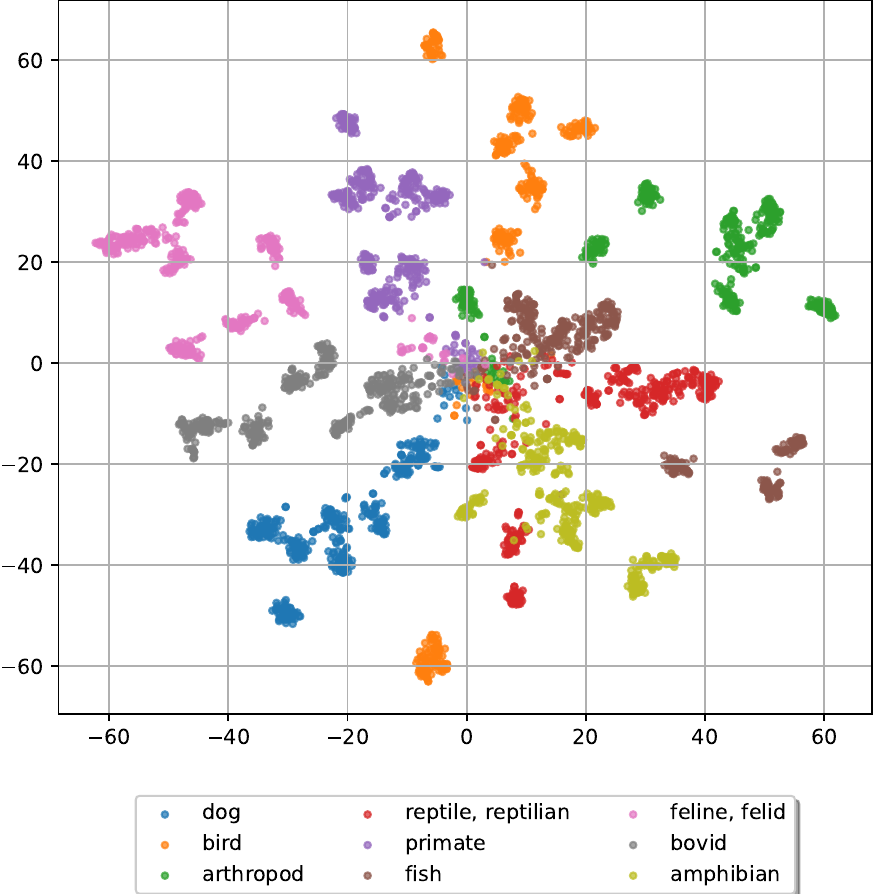}
\caption{\xclr{}}
\end{subfigure}
\hspace{0.01\linewidth}%
\begin{subfigure}[t]{0.31\linewidth}
\includegraphics[width=\textwidth]{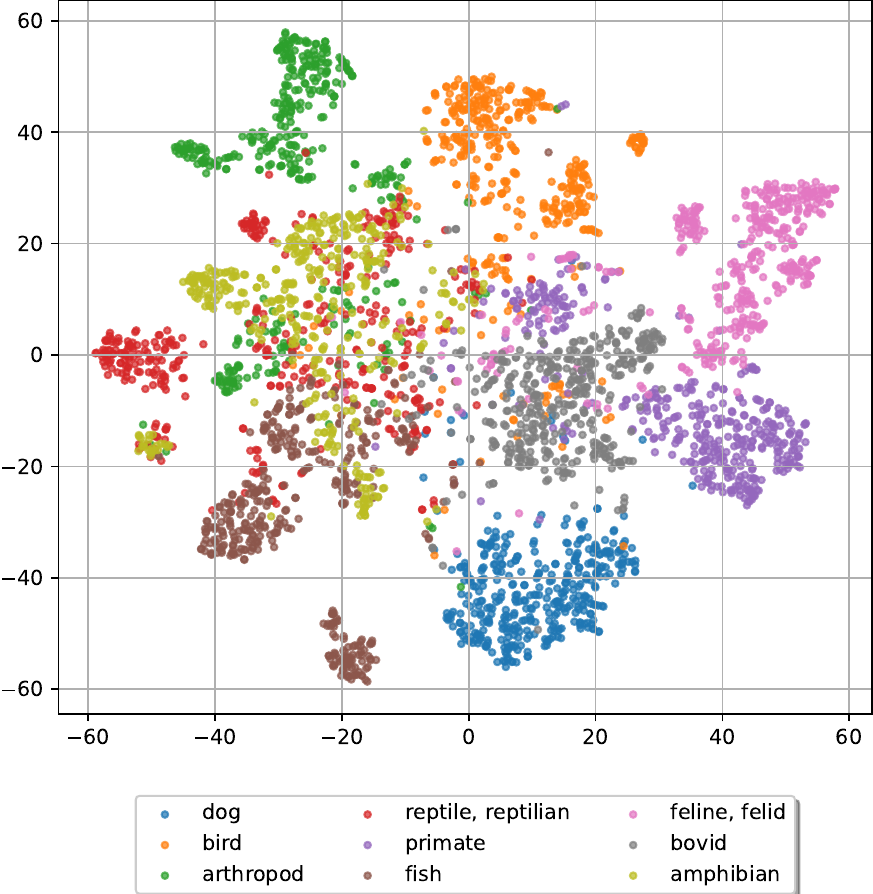}
\caption{SimCLR}
\end{subfigure}
\hspace{0.01\linewidth}%
\begin{subfigure}[t]{0.31\linewidth}
\includegraphics[width=\textwidth]{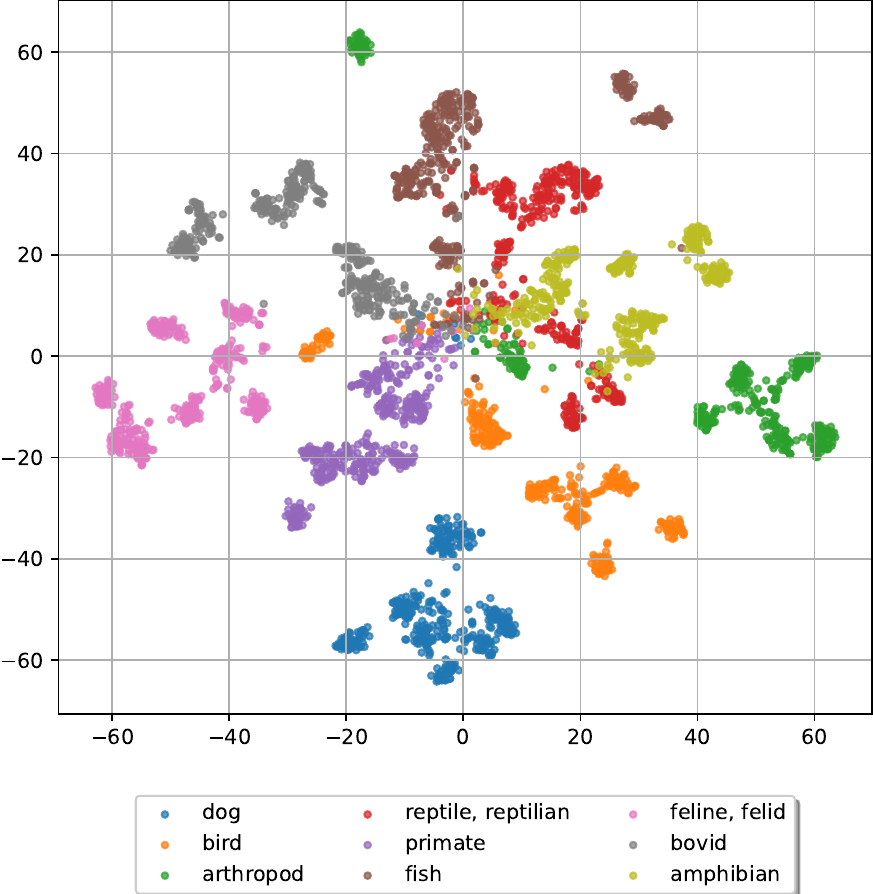}
\caption{SupCon}
\end{subfigure}
\label{fig:tsne}
\caption{\xclr{} and SupCon representations fall into a well-defined clusters, whereas SimCLR 
representations are less structured.}
\end{figure}

In \cref{fig:tsne}, we show T-SNE plots of representations of a few superclasses from ImageNet. We used the 'living 9'
set of classes from \citep{robustness}.

\end{document}